\documentclass[sigconf]{acmart}
\graphicspath{ {images/} }
\usepackage{booktabs} 
\usepackage{tikz} 
\usepackage{adjustbox} 

\setcopyright{rightsretained}

\acmDOI{10.475/123_4}

\acmISBN{123-4567-24-567/08/06}

\acmArticle{4}
\acmPrice{15.00}

\begin{document}
\title{Abstractive Tabular Dataset Summarization via Knowledge Base Semantic Embeddings}

\author{Paul Azunre, Craig Corcoran, David Sullivan, Garrett Honke, Rebecca Ruppel, Sandeep Verma, Jonathon Morgan} 
    \affiliation{ 
      \institution{New Knowledge}
    }
    \email{\string{first.name\string}@newknowledge.io}

\renewcommand{\shortauthors}{P. Azunre et al.}

\begin{abstract}
This paper describes an abstractive summarization method\footnote{Our code is available for download at https://github.com/NewKnowledge/duke} for tabular data which employs a knowledge base semantic embedding to generate the summary. Assuming the dataset contains descriptive text in headers, columns and/or some augmenting metadata, the system employs the embedding to recommend a subject/type for each text segment. Recommendations are aggregated into a small collection of \emph{super types} considered to be descriptive of the dataset by exploiting the hierarchy of types in a prespecified ontology. Using February 2015 Wikipedia as the knowledge base, and a corresponding DBpedia ontology as types, we present experimental results on open data taken from several sources -- OpenML, CKAN and data.world -- to illustrate the effectiveness of the approach.
\end{abstract}

%
%



\maketitle

\pagestyle{empty} 

\section{Introduction}

The motivation of this work is to develop a method for summarizing the content of tabular datasets. One can imagine the potential utility of automatically assigning a set of tags to each member of a large collection of datasets that would indicate the potential subject being addressed by the dataset. This can allow for semantic querying over the dataset collection to extract all available data pertinent to some specific task subject at scale.

We make the assumption that the dataset contains some text that is semantically descriptive of the dataset subject, whether appearing in columns, headers or some augmenting metadata. As opposed to an \emph{extractive} approach that would merely select some exact words and phrases from the available text, we propose an \emph{abstractive} approach that builds an internal semantic representation and produces subject tags that may not be explicitly present in the text augmenting the dataset.

The result of this work is \emph{DUKE}---{\bf D}ataset {\bf U}nderstanding via {\bf K}nowledge-base {\bf E}mbeddings---a method that employs a pretrained Knowledge Base (KB) semantic embedding to perform \emph{type recommendation} within a prespecified \emph{ontology}. This is achieved by aggregating the recommended types into a small collection of \emph{super types} predicted to be descriptive of the dataset by exploiting the hierarchical structure of the various types in the ontology. Effectively, the method represents employing an existing KB embedding to extensionally generate a \emph{dataset2vec} embedding. Using a February 2015 Wikipedia knowledge base and a corresponding DBpedia ontology to specify types, we present experimental results on open data taken from several sources---OpenML, CKAN, and data.world---to illustrate the effectiveness of the approach.

\section{Related Work}
The \emph{distributional semantics} \cite{DistributionalHypothesis} concept has been recently widely employed as a natural language processing (NLP) tool to embed various NLP concepts into vector spaces. This rather intuitive hypothesis states that the meaning of a word is determined by its context. By far the most pervasive application of the hypothesis has been the word2vec model \cite{word2vec}\cite{glove2014} which employs neural networks on large corpora to embed words that are contextually similar to be close to each other in a high-dimensional vector space. Arithmetic operations on the elements of the vector space produce semantically meaningful results, e.g., \emph{King-Man+Woman=Queen}.

Since the original word2vec model, various incremental incarnations of it have been employed to embed sentences, paragraphs and even knowledge graphs into vector spaces via sent2vec\cite{sent2vec}, paragraph2vec\cite{para2vec}, and RDF2Vec\cite{RDF2Vec} respectively.

A topic \emph{domain} is typically expressed as a manually curated ontology. A basic element of an ontology is a \emph{type}, and a \emph{type assertion} statement links specific entities of the knowledge graph to specific types.  These statements can be used to augment a semantic embedding space with type information in order to add high level context of the graph to the embedding space. For instance, it was recently shown that one can extend a pretrained Knowledge Graph Embedding (KGE) to contain types of a specific ontology if those were not already present as entities, given a list of assertion statements\cite{ISI_paper}. Thus, it can be assumed that a semantic embedding is \emph{typed} for our purposes.

We note that the abstractive tabular dataset summarization problem is closely related to the well-studied problem of \emph{type recommendation}, where the type is a super tag for all text segments in the dataset within a prespecified ontology that needs to be predicted. Systems for type recommendation using both manually curated features\cite{TYPifier} and automated features \cite{EntityTyping}, e.g., via typed KGEs\cite{ISI_paper}, for \emph{individual entities}, have been previously explored. To the best of our knowledge, this is the first application of typed semantic embeddings to abstractive tabular dataset summarization.

\section{Approach}

\subsection{Framework}

In this subsection, we present a pair of definitions to aid orientation.

\textbf{Definition (word2vec)} Word2vec models utilize a large corpus of documents to build a vector space mapping words to points in a space, where proximity implies semantic similarity\cite{word2vec}. This allows us to calculate distances between words in the dataset and the set of types in our ontology.

\textbf{Definition (wiki2vec)} When discussed in this paper, a \emph{wiki2vec} model is a form of word2vec model trained on a corpus of Wikipedia KB documents\footnote{See https://github.com/idio/wiki2vec}. Training on this data ensures that the list of types in the DBpedia ontology are included in the vocabulary of the model, and increases the likelihood that topics are discussed in context with their super-types.

Note that wiki2vec is different from a KGE, which is typically trained on relationship triples between entities in a knowledge graph (such as DBpedia)\cite{RDF2Vec}.

\subsection{Generating Type Recommendations}
The method for summarizing a tabular dataset can be broken down into three distinct steps:
\begin{enumerate}
    \item Collect a set of \emph{types} and an \emph{ontology} to use for abstraction
    \item Extract any text data from the tabular dataset and embed it into a vector space to calculate the distance to all the types in our ontology
    \item Aggregate the distance vectors for every keyword in the dataset into a single vector of distances
\end{enumerate}

\subsubsection{Type Ontology}
In order to generate an abstract term to describe the dataset, we must first collect an ontology of types to select a descriptive term from. We use an ontology provided by DBpedia\footnote{Downloaded from http://downloads.dbpedia.org/2015-10/dbpedia 2015-10.nt} which contains approximately 400 defined types, including everything from \emph{sound} to \emph{video game} and \emph{historic place}. DBpedia also contains defined parent-child relationships for the types\footnote{Defined parent-child type relationships can be found at http://dbpedia.org/ontology/} that we use to build a complete hierarchy of types e.g. that \emph{tree} is a sub-type of \emph{plant} which is a sub-type of \emph{eukaryote}.

\subsubsection{Word Embedding}
With the set of topics collected, extract each word from the dataset, embed it in a wiki2vec vector space and calculate the distance between that word and every type in the ontology. If a single cell in a column contains more than one word, take the average of the corresponding embedded vectors. This results in a collection of distance vectors representing all text in the dataset. Collect the vectors according to their source within the dataset, i.e. words in the same column are collected into a matrix of distances \emph{for each column}. If column headers are provided, treat them as an additional column in the dataset.

\subsubsection{Distance Aggregation}

The previous steps produce a set of matrices containing distances between every text segment in the dataset and the set of types. The goal of this step is to reduce them to a single vector of distances.

We utilize three successive aggregations in order to compute this final vector. The first aggregation is computed across the rows of each column matrix in order to produce a single vector of distances between the column and all types. Potential functions to use are discussed below. The second aggregation is what we call the \emph{tree} aggregation, where we take this vector of distances for a column and utilize the hierarchy of types described by DBpedia in order to update the scores for each type. For instance, we need to update the score for \emph{means of transportation} based on the scores for \emph{airplane}, \emph{train}, and \emph{automobile}. The third aggregation is performed over the set of distance vectors computed for each column, producing a single vector of distances to every defined type. 
We tested two simple functions for each aggregation step: \emph{mean} and \emph{max}, as well as a variety of more complex aggregations for the tree aggregation step. Tree aggregation allows for additional complexity because the updated distance for each type was dependent on the original distance for that type and the vector of scores for all the children. We found that the most successful tree aggregation functions were those that utilized different functions for processing the child scores and the original type score,  e.g., 
\begin{equation}
    type'= mean(type, max(child_1, child_2, ... , child_n))
    \label{eq:tree_agg}
\end{equation}

\subsubsection{Aggregation Function Selection}
\begin{figure}
    \centering
    \resizebox{0.8\columnwidth}{!}{
\begin{tikzpicture}[x=1pt,y=1pt]
\definecolor{fillColor}{RGB}{255,255,255}
\path[use as bounding box,fill=fillColor,fill opacity=0.00] (0,0) rectangle (505.89,505.89);
\begin{scope}
\path[clip] (  0.00,  0.00) rectangle (505.89,505.89);
\definecolor{drawColor}{RGB}{255,255,255}
\definecolor{fillColor}{RGB}{255,255,255}

\path[draw=drawColor,line width= 0.6pt,line join=round,line cap=round,fill=fillColor] (  0.00,  0.00) rectangle (505.89,505.89);
\end{scope}
\begin{scope}
\path[clip] ( 57.11, 70.35) rectangle (500.39,500.39);
\definecolor{drawColor}{RGB}{255,255,255}

\path[draw=drawColor,line width= 0.3pt,line join=round] ( 57.11, 89.90) --
	(500.39, 89.90);

\path[draw=drawColor,line width= 0.3pt,line join=round] ( 57.11,133.68) --
	(500.39,133.68);

\path[draw=drawColor,line width= 0.3pt,line join=round] ( 57.11,221.24) --
	(500.39,221.24);

\path[draw=drawColor,line width= 0.3pt,line join=round] ( 57.11,308.79) --
	(500.39,308.79);

\path[draw=drawColor,line width= 0.3pt,line join=round] ( 57.11,396.35) --
	(500.39,396.35);

\path[draw=drawColor,line width= 0.3pt,line join=round] ( 57.11,483.91) --
	(500.39,483.91);
\definecolor{drawColor}{RGB}{190,190,190}

\path[draw=drawColor,line width= 0.6pt,line join=round] ( 57.11,177.46) --
	(500.39,177.46);

\path[draw=drawColor,line width= 0.6pt,line join=round] ( 57.11,265.01) --
	(500.39,265.01);

\path[draw=drawColor,line width= 0.6pt,line join=round] ( 57.11,352.57) --
	(500.39,352.57);

\path[draw=drawColor,line width= 0.6pt,line join=round] ( 57.11,440.13) --
	(500.39,440.13);
\definecolor{drawColor}{RGB}{255,255,255}

\path[draw=drawColor,line width= 0.6pt,line join=round] ( 89.54, 70.35) --
	( 89.54,500.39);

\path[draw=drawColor,line width= 0.6pt,line join=round] (143.60, 70.35) --
	(143.60,500.39);

\path[draw=drawColor,line width= 0.6pt,line join=round] (197.66, 70.35) --
	(197.66,500.39);

\path[draw=drawColor,line width= 0.6pt,line join=round] (251.72, 70.35) --
	(251.72,500.39);

\path[draw=drawColor,line width= 0.6pt,line join=round] (305.78, 70.35) --
	(305.78,500.39);

\path[draw=drawColor,line width= 0.6pt,line join=round] (359.84, 70.35) --
	(359.84,500.39);

\path[draw=drawColor,line width= 0.6pt,line join=round] (413.90, 70.35) --
	(413.90,500.39);

\path[draw=drawColor,line width= 0.6pt,line join=round] (467.95, 70.35) --
	(467.95,500.39);
\definecolor{fillColor}{gray}{0.35}

\path[fill=fillColor] ( 65.22, 89.90) rectangle (113.87,480.84);

\path[fill=fillColor] (119.27, 89.90) rectangle (167.93,370.96);

\path[fill=fillColor] (173.33, 89.90) rectangle (221.99,357.83);

\path[fill=fillColor] (227.39, 89.90) rectangle (276.05,346.88);

\path[fill=fillColor] (281.45, 89.90) rectangle (330.10,314.92);

\path[fill=fillColor] (335.51, 89.90) rectangle (384.16,314.05);

\path[fill=fillColor] (389.57, 89.90) rectangle (438.22,292.59);

\path[fill=fillColor] (443.63, 89.90) rectangle (492.28,290.41);
\end{scope}
\begin{scope}
\path[clip] (  0.00,  0.00) rectangle (505.89,505.89);
\definecolor{drawColor}{RGB}{0,0,0}

\path[draw=drawColor,line width= 0.6pt,line join=round] ( 57.11, 70.35) --
	( 57.11,500.39);
\end{scope}
\begin{scope}
\path[clip] (  0.00,  0.00) rectangle (505.89,505.89);
\definecolor{drawColor}{gray}{0.30}

\node[text=drawColor,anchor=base east,inner sep=0pt, outer sep=0pt, scale=  1.40] at ( 52.16,172.64) {0.02};

\node[text=drawColor,anchor=base east,inner sep=0pt, outer sep=0pt, scale=  1.40] at ( 52.16,260.19) {0.04};

\node[text=drawColor,anchor=base east,inner sep=0pt, outer sep=0pt, scale=  1.40] at ( 52.16,347.75) {0.06};

\node[text=drawColor,anchor=base east,inner sep=0pt, outer sep=0pt, scale=  1.40] at ( 52.16,435.31) {0.08};
\end{scope}
\begin{scope}
\path[clip] (  0.00,  0.00) rectangle (505.89,505.89);
\definecolor{drawColor}{gray}{0.20}

\path[draw=drawColor,line width= 0.6pt,line join=round] ( 54.36,177.46) --
	( 57.11,177.46);

\path[draw=drawColor,line width= 0.6pt,line join=round] ( 54.36,265.01) --
	( 57.11,265.01);

\path[draw=drawColor,line width= 0.6pt,line join=round] ( 54.36,352.57) --
	( 57.11,352.57);

\path[draw=drawColor,line width= 0.6pt,line join=round] ( 54.36,440.13) --
	( 57.11,440.13);
\end{scope}
\begin{scope}
\path[clip] (  0.00,  0.00) rectangle (505.89,505.89);
\definecolor{drawColor}{RGB}{0,0,0}

\path[draw=drawColor,line width= 0.6pt,line join=round] ( 57.11, 70.35) --
	(500.39, 70.35);
\end{scope}
\begin{scope}
\path[clip] (  0.00,  0.00) rectangle (505.89,505.89);
\definecolor{drawColor}{gray}{0.20}

\path[draw=drawColor,line width= 0.6pt,line join=round] ( 89.54, 67.60) --
	( 89.54, 70.35);

\path[draw=drawColor,line width= 0.6pt,line join=round] (143.60, 67.60) --
	(143.60, 70.35);

\path[draw=drawColor,line width= 0.6pt,line join=round] (197.66, 67.60) --
	(197.66, 70.35);

\path[draw=drawColor,line width= 0.6pt,line join=round] (251.72, 67.60) --
	(251.72, 70.35);

\path[draw=drawColor,line width= 0.6pt,line join=round] (305.78, 67.60) --
	(305.78, 70.35);

\path[draw=drawColor,line width= 0.6pt,line join=round] (359.84, 67.60) --
	(359.84, 70.35);

\path[draw=drawColor,line width= 0.6pt,line join=round] (413.90, 67.60) --
	(413.90, 70.35);

\path[draw=drawColor,line width= 0.6pt,line join=round] (467.95, 67.60) --
	(467.95, 70.35);
\end{scope}
\begin{scope}
\path[clip] (  0.00,  0.00) rectangle (505.89,505.89);
\definecolor{drawColor}{gray}{0.30}

\node[text=drawColor,anchor=base,inner sep=0pt, outer sep=0pt, scale=  1.20] at ( 89.54, 57.14) {Mean,};

\node[text=drawColor,anchor=base,inner sep=0pt, outer sep=0pt, scale=  1.20] at ( 89.54, 44.18) {MeanMax,};

\node[text=drawColor,anchor=base,inner sep=0pt, outer sep=0pt, scale=  1.20] at ( 89.54, 31.22) {Mean};

\node[text=drawColor,anchor=base,inner sep=0pt, outer sep=0pt, scale=  1.20] at (143.60, 57.14) {Max,};

\node[text=drawColor,anchor=base,inner sep=0pt, outer sep=0pt, scale=  1.20] at (143.60, 44.18) {MeanMax,};

\node[text=drawColor,anchor=base,inner sep=0pt, outer sep=0pt, scale=  1.20] at (143.60, 31.22) {Mean};

\node[text=drawColor,anchor=base,inner sep=0pt, outer sep=0pt, scale=  1.20] at (197.66, 57.14) {Mean,};

\node[text=drawColor,anchor=base,inner sep=0pt, outer sep=0pt, scale=  1.20] at (197.66, 44.18) {Max,};

\node[text=drawColor,anchor=base,inner sep=0pt, outer sep=0pt, scale=  1.20] at (197.66, 31.22) {Mean};

\node[text=drawColor,anchor=base,inner sep=0pt, outer sep=0pt, scale=  1.20] at (251.72, 57.14) {Max,};

\node[text=drawColor,anchor=base,inner sep=0pt, outer sep=0pt, scale=  1.20] at (251.72, 44.18) {Max,};

\node[text=drawColor,anchor=base,inner sep=0pt, outer sep=0pt, scale=  1.20] at (251.72, 31.22) {Mean};

\node[text=drawColor,anchor=base,inner sep=0pt, outer sep=0pt, scale=  1.20] at (305.78, 57.14) {Mean,};

\node[text=drawColor,anchor=base,inner sep=0pt, outer sep=0pt, scale=  1.20] at (305.78, 44.18) {MeanMax,};

\node[text=drawColor,anchor=base,inner sep=0pt, outer sep=0pt, scale=  1.20] at (305.78, 31.22) {Max};

\node[text=drawColor,anchor=base,inner sep=0pt, outer sep=0pt, scale=  1.20] at (359.84, 57.14) {Mean,};

\node[text=drawColor,anchor=base,inner sep=0pt, outer sep=0pt, scale=  1.20] at (359.84, 44.18) {Max,};

\node[text=drawColor,anchor=base,inner sep=0pt, outer sep=0pt, scale=  1.20] at (359.84, 31.22) {Max};

\node[text=drawColor,anchor=base,inner sep=0pt, outer sep=0pt, scale=  1.20] at (413.90, 57.14) {Max,};

\node[text=drawColor,anchor=base,inner sep=0pt, outer sep=0pt, scale=  1.20] at (413.90, 44.18) {Max,};

\node[text=drawColor,anchor=base,inner sep=0pt, outer sep=0pt, scale=  1.20] at (413.90, 31.22) {Max};

\node[text=drawColor,anchor=base,inner sep=0pt, outer sep=0pt, scale=  1.20] at (467.95, 57.14) {Max,};

\node[text=drawColor,anchor=base,inner sep=0pt, outer sep=0pt, scale=  1.20] at (467.95, 44.18) {MeanMax,};

\node[text=drawColor,anchor=base,inner sep=0pt, outer sep=0pt, scale=  1.20] at (467.95, 31.22) {Max};
\end{scope}
\begin{scope}
\path[clip] (  0.00,  0.00) rectangle (505.89,505.89);
\definecolor{drawColor}{RGB}{0,0,0}

\node[text=drawColor,anchor=base,inner sep=0pt, outer sep=0pt, scale=  2.00] at (278.75,  7.44) {Model Configurations};
\end{scope}
\begin{scope}
\path[clip] (  0.00,  0.00) rectangle (505.89,505.89);
\definecolor{drawColor}{RGB}{0,0,0}

\node[text=drawColor,rotate= 90.00,anchor=base,inner sep=0pt, outer sep=0pt, scale=  2.00] at ( 19.27,285.37) {Positive Match Rate (Keep 3)};
\end{scope}
\end{tikzpicture}}
    \caption{Match rate between true labels and top 3 predicted labels for the best performing aggregation function combinations. The labels for each bar describe the three tested aggregation functions in the order: column, tree, dataset.}
    \label{fig:match_rate}
\end{figure}
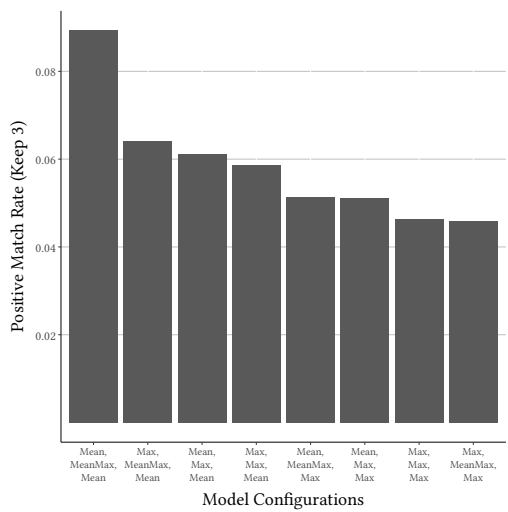

To select the best function for each aggregation, we hand-labelled a collection of datasets with types from our ontology to use as a sort of `training set'. Then, for each labelled dataset and each combination of aggregation functions, we computed the percentage of true labels found in the top three labels predicted by DUKE, with results shown in Figure \ref{fig:match_rate}. This figure clearly shows that using \emph{mean} for column aggregation, \emph{meanmax} tree aggregation described in equation \ref{eq:tree_agg}, and then \emph{mean} for the final dataset aggregation step produces the best results. 

\section{Results and Discussion}
The goal of this section is to illustrate the effectiveness of the proposed approach to the tabular dataset summarization problem, in the context of some widely available open data sets for which manually curated summary (i.e., types/tags) are available to facilitate comparison and evaluation. Links to every dataset used is provided to facilitate verification by the reader. For each dataset, we generated one subject tag using the DUKE program, as described in the previous sections, and grade it manually using `low' for low accuracy, `medium' for medium accuracy (where the automatically generated tag is ``related to'', but is not exactly one of the manual tags) and `high' for high accuracy (where the automatically generated tag is \emph{exact} in the sense that it is one of manually generated tags). Also, please note that each prediction took roughly 20 seconds to perform (approximately 17 seconds of which was spent loading the wiki2vec model) on a 16 CPU 64 GB D16s v3 Azure Cloud VM executing serially.

\subsection{Example 1 - CKAN Datasets}
Four randomly selected CKAN datasets were used: Class Size 2016-2017\footnote{Available at https://catalogue.data.gov.bc.ca/dataset/bc-schools-class-size},
2016 Annual Survey Questions\footnote{See https://catalogue.data.gov.bc.ca/dataset/bc-public-libraries-statistics-2002-2016},
BC Liquor Store Product Price List Oct 2017\footnote{Available at https://catalogue.data.gov.bc.ca/dataset/bc-liquor-store-product-price-list-historical-prices},
and Coalfile Report\footnote{Available at https://catalogue.data.gov.bc.ca/dataset/coalfile-database}. Manually curated subject tags were available for each dataset (see Table \ref{CKAN_results}). The match between the predicted tags and the manual tags for each dataset is depicted in Table \ref{CKAN_results}.

For the first two datasets, DUKE predicts an exact tag. For the next two datasets, the accuracy is medium, with \emph{wine region} being very close to \emph{wine} and \emph{river} being a common semantic theme in coal field names (examples include \emph{Elk River}, \emph{Hat Creek} and \emph{Peace River}). Specifically, the top 5 tags returned by DUKE in decreasing order for the fourth example were \emph{river}, \emph{stream}, \emph{body of water}, \emph{natural place} and \emph{natural region}, words that are semantically descriptive of the kind of names typically possessed by coal fields. Moreover, we plot the top 5 DUKE-predicted tags and the manual tags for the third example in Figure \ref{fig:embedding}, demonstrating an exact match.

\begin{table}
  \caption{CKAN tabular dataset summarization results}
  \label{CKAN_results}
  \begin{adjustbox}{width = 0.8\columnwidth, center}
  \begin{tabular}{cccl}
    \toprule
    Dataset&Manual Tags&Predicted Tag&Score\\
    \midrule
    Class Size&class size,&school&high\\
2016-2017&public, school,&&\\
&students in classes&&\\
&&&\\
2016 Annual&annual survey,&library&high\\
Survey Questions&library, public library,&&\\
&public library&&\\
&&&\\
BC Liquor Store&BC Liquor Stores,&wine region&medium\\
Product Price&alcohol, beer, price,&&\\
List Oct 2017&beverage, wine, spirits&&\\
&&&\\
Coalfile&assessment reports,&river&medium\\
Report&coal, data, maps&&\\
  \bottomrule
\end{tabular}
\end{adjustbox}
\end{table}

\subsection{Example 2 - OpenML Datasets}
Four simple OpenML datasets were obtained through the D3M DARPA program: the 185 baseball\footnote{Available for download at http://www.openml.org/d/185}, 196 autoMpg\footnote{Available for download at http://www.openml.org/d/196}, 30 personae\footnote{Available for download at http://www.clips.ua.ac.be/datasets/personae-corpus}, and 313 spectrometer\footnote{Available for download at http://www.openml.org/d/313} datasets. The results for these datasets are shown in Table \ref{OpenML_results}.

For the first two datasets, DUKE predicts an exact tag. Note that for the second dataset, we consider \emph{engine} to be an exact tag, since the manual tags are essentially attributes of engines. For the next two datasets, the accuracy is medium, with \emph{person} being very close to \emph{personality} and \emph{band} being descriptive of \emph{red band} and \emph{blue band} manual tags. To verify that bands here referred to the right context, we looked at the top 5 tags returned by DUKE, which in decreasing order are \emph{band}, \emph{brown dwarf}, \emph{inhabitants per square kilometer}, \emph{star} and \emph{celestial body}, words that are fairly consistent with the context suggested by the manual tags. Moreover, we plot these 5 tags and the manual tags for this dataset in Figure \ref{fig:embedding}, demonstrating that while an exact match is not attained, nontrivial subsets of both tag types are `very close' to each other in the wiki2vec embedding space.

\begin{table}
  \caption{OpenML tabular dataset summarization results}
  \label{OpenML_results}
  \begin{adjustbox}{width = 0.8\columnwidth, center}
  \begin{tabular}{cccl}
    \toprule
    Dataset&Manual Tags&Predicted Tag&Score\\
    \midrule
    185 baseball&baseball player,&baseball player&high\\
&play statistics,&&\\
&database&&\\
&&&\\
196 autoMpg&city-cycle,&engine&high\\
&miles per gallon,&&\\
&fuel consumption&&\\
&&&\\
30 personae&personality,&person&medium\\
&prediction,&&\\
&from text&&\\
&&&\\
313 spectrometer&measurement, sky,&band&medium\\
&red band, blue band,&&\\
&spectrum, database,&&\\
&flux&&\\
  \bottomrule
\end{tabular}
\end{adjustbox}
\end{table}

\subsection{Example 3 - data.world Datasets}
The names of some randomly-selected data.world datasets are as follows: US terrorist origins\footnote{Available for download at https://data.world/tommyblanchard/u-s-terrorist-origins}, Occupational Employment Growth\footnote{Available for download at https://data.world/tommyblanchard/u-s-terrorist-origins}, CAFOD activity file for Haiti \footnote{Available for download at https://data.world/cafod/cafod-activity-file-for-haiti} and Queensland Gambling Data \footnote{Downloaded at https://data.world/queenslandgov/all-gambling-data-queensland}. The results for this representative set of four data.world datasets are shown in Table \ref{data.world_results}.

\begin{figure}
    \centering
    \resizebox{0.99\columnwidth}{!}{\input{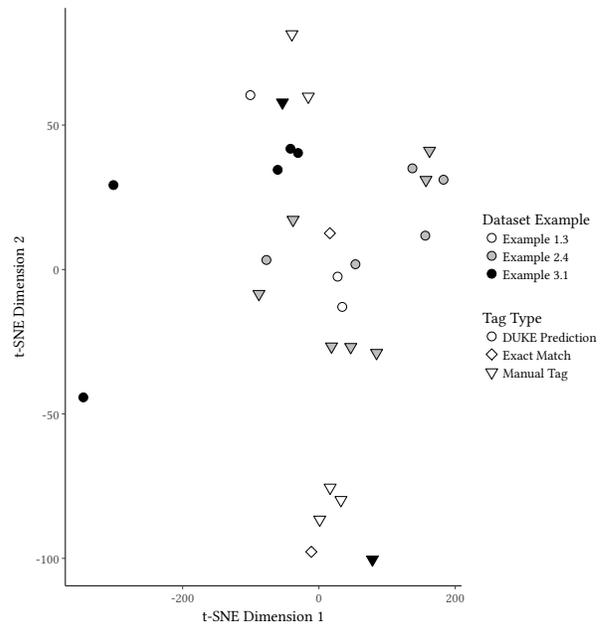}}
    \caption{Concept embedding space for three of the examined datasets. Point shape depicts DUKE predictions and manual tags. t-SNE dimension reduction was used to project the 1000 dimension concept embeddings into a 2D space for presentation.}
    \label{fig:embedding}
\end{figure}

For the first two datasets, DUKE achieves medium accuracy. To see the justification for this, note that the top 5 tags returned by DUKE for the first dataset, in decreasing order, are \emph{person}, \emph{still image}, \emph{legal case}, \emph{supreme court of the USA} and \emph{military person}, words fairly descriptive of the dataset, which is a list of terrorists, availability of a headshot and details of their legal charges. Moreover, we plot these 5 tags and the manual tags for this dataset in Figure \ref{fig:embedding}, demonstrating that while an exact match is not attained, nontrivial subsets of both tag types are `very close' to each other in the wiki2vec embedding space. The second dataset provides a list of occupations, many of which are scientific, and corresponding wages at various locations, which leads us to believe that \emph{site of scientific interest} is fairly descriptive of the semantics represented in the dataset. For the next two datasets, the accuracy is high, which should be self-explanatory to the reader from Table \ref{data.world_results}.

\begin{table}
  \caption{Tabular dataset summarization data.world results}
  \label{data.world_results}
  \begin{adjustbox}{width = 0.8\columnwidth, center}
  \begin{tabular}{cccl}
    \toprule
    Dataset&Manual Tags&Predicted Tag&Score\\
    \midrule
    US terrorist&terrorism,&person&medium\\
origins&usa politics&&\\
&&&\\
Occupational&employment,&site of&medium\\
Employment&economics&scientific&\\
Growth&&interest&\\
&&&\\
CAFOD activity&funding, Haiti,&human&high\\
file for&grants, donors,&development&\\
Haiti&aid transparency&index&\\
&&&\\
Queensland&expenditure,&casino&high\\
Gambling&gambling,&&\\
Data&Queensland&&\\
  \bottomrule
\end{tabular}
\end{adjustbox}
\end{table}

\section{Conclusion and Future Work}
A method for abstractive summarization of tabular datasets, under the assumption that it contains some descriptive text, was presented. Results of numerical experiments on OpenML, CKAN and data.world datasets show good agreement between manual and automatically generated tags by our system, DUKE. These results can be significantly improved by more extensive ontologies included in the model (in place of the 2015 DBpedia ontology). Additionally, retraining wiki2vec on a more complete version of DBpedia (potentially augmented using an Automatic Knowledge Base Completion, or AKBC, Algorithm \cite{AKBC1} \cite{AKBC2}) will help improve the accuracy of our system. More sophisticated handling of multi-word phrases also needs to be explored.

\textbf{Acknowledgements} Work was supported by the Defense Advanced Research Projects Agency (DARPA) under Contract Number D3M (FA8750-17-C-0094). Views, opinions, and findings contained in this report are those of the authors and should not be construed as an official Department of Defense position, policy, or decision.

\bibliographystyle{ACM-Reference-Format}
\bibliography{bibliography}

\end{document}